\documentclass{article}

\usepackage{arxiv}

\usepackage[utf8]{inputenc} % allow utf-8 input
\usepackage[T1]{fontenc}    % use 8-bit T1 fonts
\usepackage{hyperref}       % hyperlinks
\usepackage{url}            % simple URL typesetting
\usepackage{booktabs}       % professional-quality tables
\usepackage{amsfonts}       % blackboard math symbols
\usepackage{nicefrac}       % compact symbols for 1/2, etc.
\usepackage{microtype}      % microtypography
\usepackage{lipsum}
\usepackage{graphicx}
\usepackage{adjustbox}
\usepackage{array}
\usepackage{float}
\graphicspath{ {./figures/} }

\title{Semantic Segmentation of Unmanned Aerial Vehicle Remote Sensing Images using SegFormer}

\author{
 Vlatko Spasev \\
  Faculty of Computer Science and Engineering\\
  University Ss Cyril and Methodius\\
  Skopje 1000, North Macedonia \\
  \texttt{vlatko.spasev@finki.ukim.mk} \\
  \And
   Ivica Dimitrovski \\
  Faculty of Computer Science and Engineering\\
  University Ss Cyril and Methodius\\
  Skopje 1000, North Macedonia \\
  \texttt{ivica.dimitrovski@finki.ukim.mk} \\
   \And
    Ivan Chorbev \\
  Faculty of Computer Science and Engineering\\
  University Ss Cyril and Methodius\\
  Skopje 1000, North Macedonia \\
  \texttt{ivan.chorbev@finki.ukim.mk} \\
   \And
 Ivan Kitanovski \\
  Faculty of Computer Science and Engineering\\
  University Ss Cyril and Methodius\\
  Skopje 1000, North Macedonia \\
  \texttt{ivan.kitanovski@finki.ukim.mk} \\
}

\begin{document}
\maketitle
\begin{abstract}
The escalating use of Unmanned Aerial Vehicles (UAVs) as remote sensing platforms has garnered considerable attention, proving invaluable for ground object recognition. While satellite remote sensing images face limitations in resolution and weather susceptibility, UAV remote sensing, employing low-speed unmanned aircraft, offers enhanced object resolution and agility. The advent of advanced machine learning techniques has propelled significant strides in image analysis, particularly in semantic segmentation for UAV remote sensing images. This paper evaluates the effectiveness and efficiency of SegFormer, a semantic segmentation framework, for the semantic segmentation of UAV images. SegFormer variants, ranging from real-time (B0) to high-performance (B5) models, are assessed using the UAVid dataset tailored for semantic segmentation tasks. The research details the architecture and training procedures specific to SegFormer in the context of UAV semantic segmentation. Experimental results showcase the model's performance on benchmark dataset, highlighting its ability to accurately delineate objects and land cover features in diverse UAV scenarios, leading to both high efficiency and performance.
\end{abstract}
\section{Introduction}
In recent years, the increasing utilization of Unmanned Aerial Vehicles (UAVs) as remote sensing platforms has generated substantial interest and served as valuable resources for ground object recognition \cite{Osco_2021}. Satellite remote sensing images, while widely used, often exhibit limitations such as low resolution for low-altitude targets and susceptibility to weather conditions, leading to obscured ground objects and challenges in object recognition. In contrast, UAV remote sensing employs low-speed unmanned aircraft as aerial platforms equipped with infrared and camera technology to capture image data. UAVs, flying at lower altitudes compared to satellites, can closely approach the ground, enhancing object resolution significantly. The close-range image resolution can reach the centimeter level, enabling the efficient collection of low-altitude, high-resolution aerial images in a timely and cost-effective manner \cite{CHENG20241}. Camera tilt is pivotal in shaping UAV imagery quality and coverage. Vertical aerial photography, with a perpendicular camera axis, offers limited coverage. In contrast, low oblique images result from deliberate tilting (15° to 30°), excluding the horizon and providing a broader perspective. High oblique imagery, with a greater tilt (approximately 60°), captures a larger land area. The visible horizon distinguishes high oblique photos, making them suitable for comprehensive landscape analysis and documentation.

Recent advancements in machine learning, along with the wealth of remote sensing data available, have significantly improved image analysis and interpretation \cite{merdjanovska2022crop}, \cite{kemker2018algorithms}, \cite{DIMITROVSKI202318}, \cite{Dimitrovski2024}. A particularly exciting area of research is semantic segmentation for UAV remote sensing images, which allows for precise analysis of ground objects and their relationships \cite{YUAN2021114417}, \cite{Spasev_2023}. Semantic segmentation tasks focus on labeling each pixel of an image with a corresponding class of what the pixel represents. This results in a detailed segmentation map where every pixel has a specific class designation. This technique allows for fine-grained object identification within an image, unlike object detection which focuses on broader localization of objects. Semantic segmentation of remote sensing images is a fundamental task in the field of remote sensing and computer vision \cite{rs15092343}. The goal is to partition the image into meaningful regions, enabling detailed analysis and understanding of the Earth's surface. This finer granularity of analysis provides a more profound understanding of the intricate spatial distribution of features within remote sensing images. In the past few years, global research interest in UAV remote sensing has surged, driven by its mobility, speed, and economic advantages. Evolving from research and development, this technology has transitioned to practical applications, positioning itself as one of the forefront aerial remote sensing technologies for the future.  

Semantic segmentation of UAV remote sensing images finds diverse applications, including environmental monitoring \cite{green2019using}. Its ability to swiftly update, correct, and enhance geo-environmental information and outdated GIS databases provides crucial technical support for the administration of government and related departments, as well as for land and geo-environmental management. Furthermore, UAV remote sensing proves valuable in electric power inspection \cite{zhang2017automatic}, agricultural monitoring \cite{zhang2021review}, high-speed patrol \cite{yang2021programming}, disaster monitoring and prevention \cite{kamilaris2018disaster}, meteorological detection \cite{funk2017passive}, aerial survey \cite{de2017fault}, and various other applications. The effectiveness of semantic segmentation methods is paramount for the practical implementation of various applications. As these applications continue to evolve, there is a growing demand for real-time execution of semantic segmentation \cite{Safavi_2023}.

This paper assesses the effectiveness and efficiency of SegFormer \cite{xie2021segformer} in semantic segmentation tasks involving UAV images. SegFormer comes in several variants, denoted by code names B0 to B5. Among these, B0 is the smallest model tailored for real-time applications, while B5 is the largest model designed for high performance. We chose the UAVid dataset \cite{LYU2020108}, specifically curated for UAV video data in semantic segmentation tasks, with a focus on urban scenes. The evaluation of effectiveness involves reporting the mean intersection over union ($mIoU$), while efficiency is gauged through metrics such as the number of parameters, frames per second (FPS), and latency for the different SegFormer variants. Latency denotes the duration it takes for the model to analyze an image and provide information about the identified segments/objects.

The paper is organized as follows: Section 2 provides a review of existing research on semantic segmentation techniques applied to UAV remote sensing images. Section 3 elucidates the key features of the SegFormer semantic segmentation framework. Section 4 provides an overview of the dataset utilized in the research. Section 5 comprehensively outlines the experimental setup, including data preprocessing, training protocols, model parameters, and evaluation measures. Section 6 presents the experimental results alongside relevant discussions. Finally, Section 7 concludes the paper, summarizing the findings and contributions.

\section{Related Work}
Semantic segmentation extends the concept of image classification by assigning a class label to each pixel in an image, rather than just the entire image. Semantic segmentation in remote sensing images presents unique challenges due to factors such as high resolution, complex spatial structures, diverse object scales, and large data volumes. Early attempts at semantic segmentation relied on traditional machine learning methods. These methods fell into two main categories: pixel-based and region-based approaches \cite{weinland2011survey}. However, they had limitations.  They depended heavily on manually designed features and setting thresholds for those features, which could be time-consuming and ineffective for complex images. These images often have challenges like different lighting conditions, textures, and object sizes. Traditional methods often struggled with these complexities, leading to inconsistent performance and limited usefulness. This paved the way for the adoption of modern deep learning techniques \cite{YUAN2021114417}.

The advent of deep learning, particularly with the introduction of Convolutional Neural Networks (CNNs) and Fully Convolutional Networks (FCNs), has transformed the field of semantic segmentation \cite{guo2018review}. FCNs, often combined with encoder-decoder architectures, are now the leading approach. Early FCNs used repeated convolutions and pooling to make predictions for each pixel \cite{long2015fully}. Newer models, like U-Net \cite{Ronneberger2015} and SegNet \cite{badrinarayanan2017segnet}, combine high-level features (capturing large-scale information) with low-level details (preserving sharp boundaries) during decoding. This improves both capturing the overall scene and precisely identifying objects. To see more of an image at once (increasing the receptive field), techniques, like dilated convolutions were introduced in DeepLab \cite{chen2017deeplab}. Later advancements like PSPNet \cite{zhao2017pyramid} and UperNet \cite{xiao2018unified} incorporated spatial pyramid pooling to capture information at different scales within the image. DeepLabV3+ combined these ideas into a powerful yet efficient architecture \cite{chen2018encoder}. Subsequent advancements, as demonstrated by PSANet \cite{zhao2018psanet} and DRANet \cite{fu2020scene}, have replaced traditional pooling methods with attention mechanisms applied to encoder feature maps, enhancing the ability to capture long-range dependencies.

Most recently, researchers have explored using transformers, a type of neural network architecture that excels at capturing long-range relationships between image parts. Models like Segmenter \cite{strudel2021segmenter}, SegFormer \cite{xie2021segformer}, and MaskFormer \cite{cheng2021per} all leverage transformers for improved performance. Segmenter uses a specialized transformer backbone and a mask decoder, while SegFormer offers a simpler yet effective solution with transformers as encoders and lightweight decoders. Inspired by DEtection TRansformer (DETR) \cite{carion2020end}, MaskFormer uses transformers to directly generate object masks, making it versatile for various segmentation tasks. To address limitations in MaskFormer, Mask2Former \cite{cheng2021mask2former} introduced a multi-scale decoder and a masked attention mechanism.

Semantic segmentation of UAV images presents a formidable challenge due to a confluence of factors \cite{Spasev_2023}. The diverse nature of UAV imagery, characterized by a wide spectrum of resolutions and object orientations, necessitates robust models capable of generalizing across disparate datasets. The inherent scale variability within single images, ranging from expansive structures to diminutive objects, demands models that can adeptly handle such disparities, accurately segmenting both large-scale and fine-grained elements. Densely populated urban environments and the minute details often found in natural landscapes pose significant obstacles to precise object delineation. Moreover, the imbalanced distribution of classes within UAV imagery, where certain categories are represented far less frequently than others, hinders model training and can lead to biased segmentation results. Finally, the intricate and cluttered backgrounds common in aerial imagery introduce additional complexity, making the separation of objects from their surroundings a demanding task.

The complex nature of remote sensing imagery, characterized by varying resolutions, diverse object scales, and intricate background patterns, has necessitated the development of sophisticated semantic segmentation techniques \cite{yamazaki2023aerialformer}, \cite{wang2022unetformer}, \cite{he2023building}. Recent advancements in deep learning have spurred significant progress in this domain. AerialFormer, for example, presents a hierarchical framework that effectively captures multi-scale features through a Transformer encoder while refining segmentation details using a Multi-Dilated Convolutional Neural Network (MD-CNN) decoder \cite{yamazaki2023aerialformer}. UNetFormer, on the other hand, introduces a global-local Transformer block (GLTB) to enhance contextual understanding, coupled with a feature refinement head for precision \cite{wang2022unetformer}. To optimize computational efficiency, the model integrates a lightweight CNN-based encoder with a Transformer decoder. A novel approach is embodied by the Uncertainty-Aware Network (UANet), which incorporates uncertainty modeling to mitigate the challenges posed by complex background elements and improve the accuracy of building footprint segmentation \cite{he2023building}. These innovative architectures collectively demonstrate the ongoing exploration of effective strategies for tackling the unique challenges inherent in remote sensing image analysis. By leveraging the power of deep learning and addressing specific limitations, researchers are steadily advancing the field of semantic segmentation in this domain.

\section{Model Architecture}
\begin{figure}
    \centering
        \includegraphics[width=1.0\linewidth]{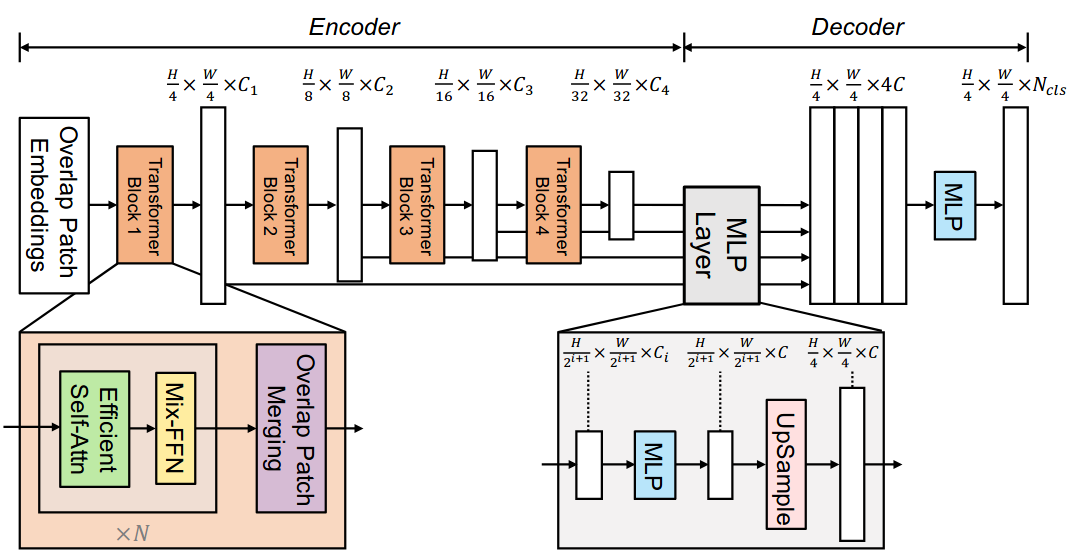}
    \caption{Illustration of the SegFormer semantic segmentation framework architecture. The image is taken from \cite{xie2021segformer}.}
    \label{fig:segformer_architecture}
\end{figure}

SegFormer is a semantic segmentation model that, unlike traditional methods heavily reliant on convolutional neural networks (CNNs), combines Transformers with lightweight multilayer perceptron (MLP) decoders \cite{xie2021segformer}. Figure~\ref{fig:segformer_architecture} illustrates the architecture of the SegFormer semantic segmentation framework.

The SegFormer model incorporates a novel hierarchically structured Transformer encoder that produces multiscale features. This component excels at capturing intricate relationships between distant image regions, crucial for tasks like segmenting objects with complex shapes or fine details. Additionally, the encoder's design allows it to extract features at various resolutions. This proves beneficial for segmentation as it captures both the overall image context and minute details necessary for accurate pixel-by-pixel classification. The encoder that is used in SegFormer is named Mix Transformer (MiT). A series of MiT encoders labeled MiT-B0 through MiT-B5, has been designed with identical architectures but differing sizes. MiT-B0 is utilized as the lightweight model for rapid inference, while MiT-B5 is employed as the largest model to achieve optimal performance.

Furthermore, SegFormer does not rely on positional encoding, which is a standard approach in transformer-based models. This technique, originally used in natural language processing, accounts for the order of words in a sentence. In image data, the relative position of pixels is inherent, making positional encoding redundant in the case of SegFormer. Not requiring positional encoding alleviates the need for interpolating positional codes, which otherwise degrades performance when the testing resolution diverges from the training resolution. 

Finally, what is also specific about SegFormer is its' lightweight decoder setup. It employs an MLP decoder that aggregates information from different layers. This design choice maintains good performance while ensuring efficiency. The decoder is key in combining information from the different encoder outputs, effectively merging local and global attention resulting in robust representations for precise segmentation.

\section{Dataset}
The UAVid dataset \cite{LYU2020108} is composed of 420 images, each with dimensions of 4096×2160 or 3840×2160 pixels. For training and validation purposes, 200 and 70 images were allocated, respectively, while the remaining images were dedicated to testing. Captured within complex urban settings from an oblique perspective at a 50-meter altitude, the UAVid images present a variety of stationary and moving objects. The dataset exhibits a side view perspective, offering a unique vantage point for analysis.

\begin{figure}[!ht]
    \centering
        \includegraphics[width=0.7\linewidth]{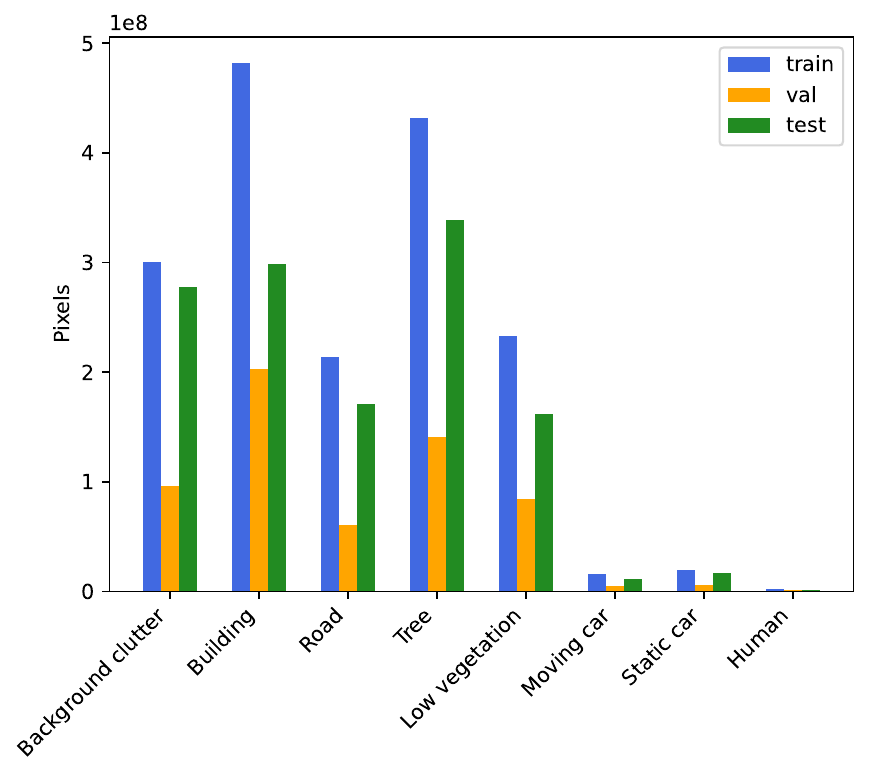}
    \caption{The pixel distribution across labels in the train, validation, and test splits of the UAVid dataset.}
    \label{fig:dataset_distribution}
\end{figure}

Comprising eight distinct classes, the UAVid dataset categorizes objects into building, road, static car, tree, low vegetation, human, moving car, and background clutter. The UAVid dataset encompasses a comprehensive set of urban scene elements, including residential and under-construction buildings collectively categorized as buildings, stationary and moving vehicles, clearly defined road surfaces (excluding parking lots and sidewalks), and a background clutter class encompassing miscellaneous urban features. Buildings, trees, and roads constitute the predominant visual components within UAVid images, while cars and pedestrians represent a smaller but significant portion of the dataset, accounting for approximately 3\% of the total class distribution. A detailed breakdown of pixel distribution across the training, validation, and testing subsets is presented in Figure~\ref{fig:dataset_distribution}. To provide visual context, Figure~\ref{fig:example_inference} offers representative examples of UAVid images alongside their corresponding ground truth masks, showcasing the dataset's rich and varied content.

\section{Experimental Setup}
Original images of UAVid dataset are very large, thus pre-processing of them is adopted. Utilizing a fixed clip size of 512 pixels and a stride size of 256 pixels for generating clipped images, the intersection of width and height is calculated to ensure comprehensive coverage of the entire image. Via this procedure, a cumulative sum of 8000 images was assigned for training, with an additional 2800 images specifically designated for validation. The images maintain a resolution of 512x512 pixels. The images within the test split remained unaltered and underwent no modifications.

We are evaluating the performance of three SegFormer variants utilizing distinct MiT encoders, namely MiX-B0, MiX-B3, and MiX-B5 as encoders with input size of 512. Additionally, the fine-tuned versions of the MiT encoders on the ImageNet-1k dataset are incorporated into our analysis \cite{xie2021segformer}. We evaluate the efficiency of the different encoders by reporting the number of parameters, FPS, and latency. Additionally, we evaluate the performance of an ensemble method, which involves combining the SegFormer-B3 and SegFormer-B5 models. In this ensemble approach, we derive the final predictions by calculating the geometric mean of the base models' predictions.

Models were trained on the training split, with hyperparameter tuning conducted using the validation set. To mitigate overfitting, early stopping was implemented, terminating training if validation loss failed to improve over 20 epochs. The optimal model, determined by the highest evaluation metric on the validation set, was retained and subsequently evaluated on the unseen test split to assess final performance. The maximum training duration was capped at 100 epochs.

To bolster model robustness and generalization on the UAVid dataset, we implemented a multifaceted data augmentation pipeline during training. This process encompassed a combination of standard and more complex image transformations. Basic augmentations included random horizontal flipping and adjustments to image brightness and contrast. To introduce additional variability and challenge the model, we randomly applied more intricate transformations such as contrast limited adaptive histogram equalization, grid distortion, and optical distortion \cite{info11020125}. To ensure consistency with the ImageNet-1k dataset, we standardized image pixel values using their corresponding mean and standard deviation. It is essential to note that these augmentations were exclusively applied during the training phase. When evaluating model performance on the validation and test sets, images underwent only normalization to provide an unbiased assessment. This rigorous data augmentation approach was instrumental in improving model resilience to variations within the UAVid dataset, ultimately enhancing overall segmentation accuracy.

Our experimental protocol adhered to a fixed batch size of 12 samples per training iteration. To optimize model parameters, we leveraged the AdamW optimizer, a variant of Adam that incorporates weight decay, with an initial learning rate set to $1e-4$ \cite{loshchilov2017decoupled}. Recognizing the potential benefits of a gradually decreasing learning rate in stabilizing training and enhancing convergence, we employed a polynomial decay schedule. This learning rate adjustment strategy involves a polynomial function that systematically reduces the learning rate from an initial value of $1e-4$ to a final value of $1e-7$ over a specified number of training steps. To achieve a robust and balanced optimization of the semantic segmentation model, we adopted a hybrid loss function that combines the strengths of Cross Entropy loss and Dice loss. Cross Entropy loss, a standard measure of classification performance, quantifies the pixel-wise discrepancy between predicted and ground truth segmentation masks. However, it may not adequately capture object boundaries, a critical aspect of semantic segmentation. To address this limitation, we incorporated Dice loss, a metric that specifically focuses on the overlap between predicted and ground truth regions. By combining these complementary loss functions, our model was able to effectively learn discriminative features, accurately localize object boundaries, and achieve a comprehensive understanding of the image content. 

To process large input images, we adopt a sliding window approach with a 1024-pixel window size and a 128-pixel overlap between adjacent patches \cite{sermanet2014overfeat}. To further enhance prediction accuracy, we incorporate test-time augmentation (TTA) by horizontally flipping input images. The final segmentation map is obtained by averaging the predictions from both the original and flipped images. To assess model performance, we employed the Intersection over Union ($IoU$) metric, calculated as the ratio of the overlapping area between predicted and ground truth segmentation masks to their combined area. Additionally, we reported the mean IoU ($mIoU$) as an aggregate performance indicator across all classes. All experiments were conducted on NVIDIA A100-PCIe GPUs with 40GB memory using CUDA 11.5. The PyTorch Lightning framework facilitated model development and training \cite{Falcon_PyTorch_Lightning_2019}.

\section{Results}
The results of the experiments are presented in Table~\ref{tab:results_uavid}, showcasing the label-wise $IoU$ in percentage, along with the mean $IoU$ for each of the models. Based on the results, it can be inferred that the SegFormer model with the MiX-B5 encoder yields slightly better results compared to the model trained with the MiX-B3 encoder. The SegFormer model with the MiX-B0 encoder exhibits the lowest prediction performance. Furthermore, the incorporation of test-time augmentation contributes to improving the results of the models. Ensemble methods play a vital role in further enhancing the predictions of base SegFormer models. In Table~\ref{tab:results_uavid}, the model labeled as \emph{Ensemble} represents a fusion of the predictions from the base SegFormer-B3 and SegFormer-B5 models, while \emph{Ensemble (tta)} denotes a fusion of the predictions obtained through test-time augmentation of the base SegFormer-B3 and SegFormer-B5 models. The model Ensemble (tta) achieves the highest $mIoU$, surpassing the performance of the base SegFormer models. In the comparison, we have included existing methods such as U-Net with ResNet50 encoder \cite{Ronneberger2015}, DeepLabV3+ with ResNet50 encoder \cite{chen2017deeplab}, Category attention guided network (CAGNet) \cite{WANG2024103661}, UNetFormer with ResNet18 encoder, Densely Connected Swin Transformer (DC-Swin) with Swin-S encoder \cite{Wangdcswin}. When looking at the results, SegFormer with the MiX-B5 encoder performs competitively against other methods, it outperforms the models included in the comparison.

\begin{table}[ht]\centering
\caption{Comparative performance analysis of SegFormer and established semantic segmentation approaches on the UAVid dataset. Performance is evaluated using Mean Intersection over Union ($mIoU \%$) and labels-wise Intersection over Union ($IoU \%$) metrics. Top-performing models for each label are indicated in bold, with the second-best performance denoted by underlining.}\label{tab:results_uavid}
\scriptsize \begin{adjustbox}{width=1.0\linewidth}
\begin{tabular}{l|>{\centering}p{0.07\textwidth} >{\centering}p{0.07\textwidth} >{\centering}p{0.07\textwidth}  >{\centering}p{0.07\textwidth}  >{\centering}p{0.07\textwidth}  >{\centering}p{0.07\textwidth}  >{\centering}p{0.07\textwidth}  >{\centering}p{0.07\textwidth} | c}\toprule
Model \textbackslash Label  &  \rotatebox{90}{Clutter}  & \rotatebox{90}{Buildings}  & \rotatebox{90}{Road}  & \rotatebox{90}{Tree}  & \rotatebox{90}{Low vegetation} & \rotatebox{90}{Moving car} & \rotatebox{90}{Static car} & \rotatebox{90}{Human} & mIoU  \\\midrule

U-Net \cite{Ronneberger2015} &  67.69 & 87.28 & 80.2 & 79.69 & 63.46 & 70.72 & 58.11 & 30.62 & 67.22 \\ 
DeepLabV3+ \cite{chen2017deeplab} &  67.86 & 87.87 & 80.23 & 79.74 & 62.03 & 71.51 & 62.99 & 29.5 & 67.72 \\ 
CAGNet \cite{WANG2024103661} & 69.8  & 88.4  & \underline{82.7}  & 80.6 & 64.6 & 76.0 & 57.8 & 32.1 & 69.0 \\ 
UNetFormer \cite{wang2022unetformer} & 68.4 & 87.4 & 81.5 & 80.2 & 63.5 & 73.6 & 56.4 & 31.0 & 67.8 \\ 
DC-Swin \cite{Wangdcswin} &  \underline{70.72} & \textbf{89.66} & \textbf{83.42} & 80.75 & 65.23 & 74.97 & 59.77 & 32.02 & 69.57  \\ 
\midrule
SegFormer-B0 & 65.55 & 85.97 & 78.31 & 79.3 & 62.94 & 70.05 & 58.4 & 28.99 & 66.19  \\
SegFormer-B0 (tta)  &  66.37 & 86.57 & 79.16 & 79.8 & 63.5 & 71.25 & 58.66 & 29.94 & 66.91 \\
SegFormer-B3  & 68.8  & 88.46 & 80.19 & 80.54 & 65.24 & 73.44 & 64.92 & 32.21 & 69.22 \\
SegFormer-B3 (tta)  & 69.46 & 88.81 & 80.77 & 81.03 & 65.88 & 73.75 & 65.88 & 32.81 & 69.8  \\
SegFormer-B5  &  69.69  &  88.08  &  82.15  &  80.42  &  63.96  & 75.12
& 65.16 & 31.83 & 69.55 \\
SegFormer-B5 (tta)  &  70.21  &  88.41  &  82.54  &  80.81  &  64.54  & \textbf{76.38} & \underline{66.31} & 32.61 &   70.23  \\
Ensemble  &  70.33 & 88.9 & 81.98 & \underline{81.24} & \underline{65.94} & 75.55 & 66.18 & \underline{32.83} & \underline{70.37} \\
Ensemble (tta)  &  \textbf{70.85} & \underline{89.19} & 82.46 & \textbf{81.57} & \textbf{66.26} & \underline{76.17} & \textbf{67.39} & \textbf{33.16} & \textbf{70.88}  \\
\bottomrule
\end{tabular} \end{adjustbox}
\end{table}

\begin{figure}[h]
    \centering
        \includegraphics[width=0.9\linewidth]{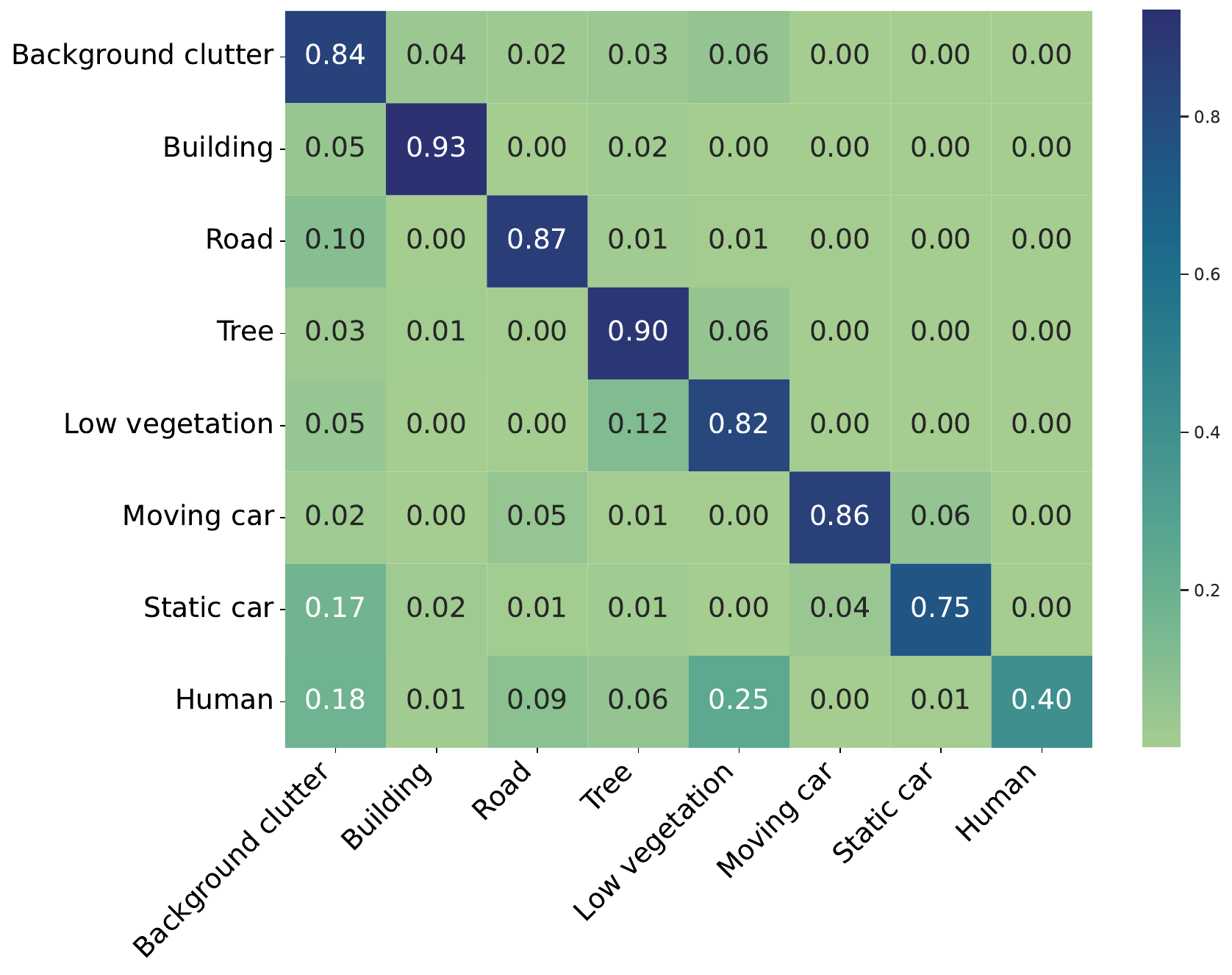}
    \caption{Confusion matrix obtained from the Ensemble (tta) model as in Table~\ref{tab:results_uavid}.}
    \label{fig:confmat}
\end{figure}

\begin{figure}
    \centering
        \includegraphics[width=0.9\linewidth]{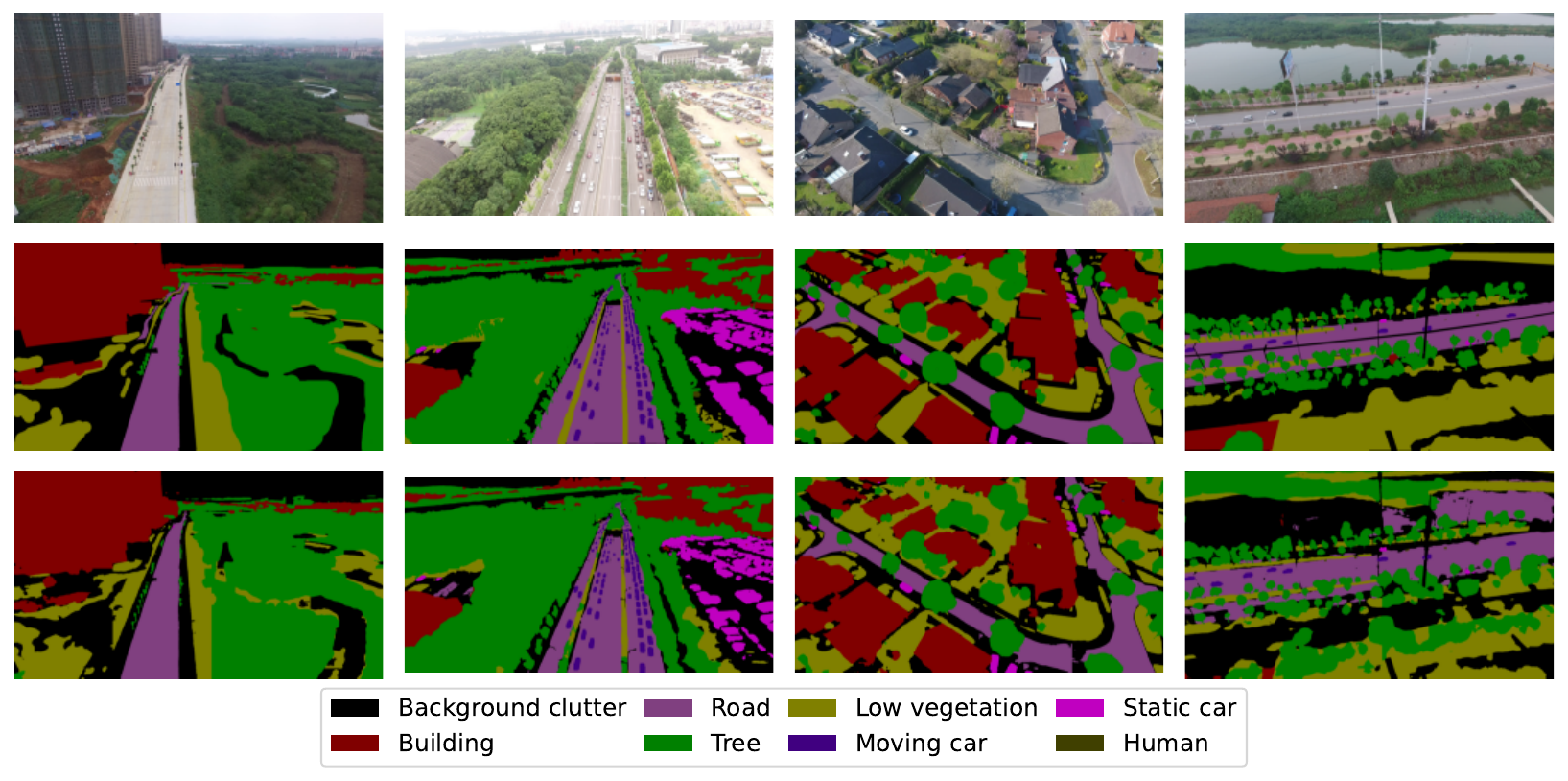}
    \caption{Example images and corresponding ground truth and predicted masks from the UAVid dataset. The first row presents UAV-captured images, while the second row displays their respective ground truth segmentation masks. The third row showcases the segmentation results produced by the Ensemble (tta) model as detailed in Table~\ref{tab:results_uavid}.}
    \label{fig:example_inference}
\end{figure}

\begin{figure}
    \centering
        \includegraphics[width=0.9\linewidth]{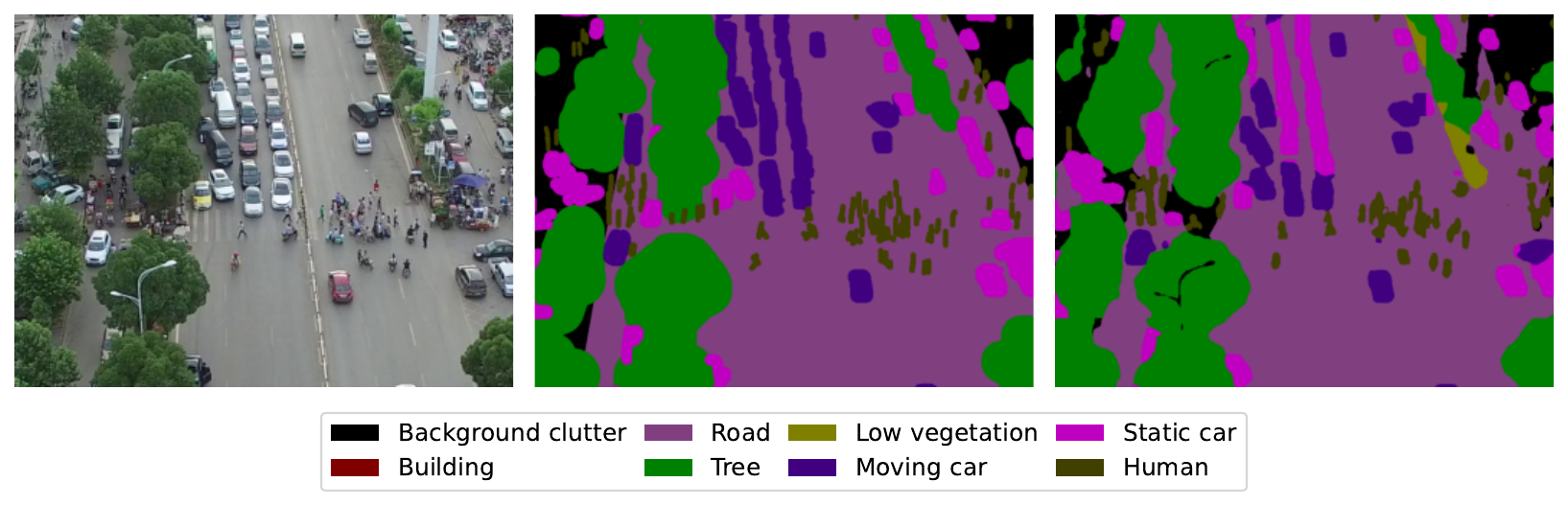}
    \caption{Zoomed-in view of a complex urban scene highlighting pedestrians and moving vehicles from the UAVid dataset, ground truth mask, and the predicted mask obtained using the Ensemble (tta) model, as outlined in Table~\ref{tab:results_uavid}.}
    \label{fig:example_inference_1}
\end{figure}

Figure~\ref{fig:example_inference} offers a visual representation of the UAVid dataset, showcasing sample images, their corresponding ground truth masks, and the segmentation outputs generated by our proposed model. The quantitative evaluation metrics presented in Table~\ref{tab:results_uavid}, specifically the Intersection over Union (IoU) scores for individual labels, highlight the model's strengths and weaknesses. Notably, the model exhibits superior performance in segmenting buildings, roads, trees, and moving vehicles. Conversely, the 'Human' label poses a significant challenge, resulting in considerably lower IoU values. A deeper examination of the confusion matrix presented in Figure~\ref{fig:confmat} provides valuable insights into the model's error patterns. The recurrent misclassifications between semantically similar classes such as 'Moving car' and 'Static car', as well as 'Tree' and 'Low vegetation', are unsurprising. The confusion between 'Human' and 'Low vegetation' is particularly noteworthy, likely attributed to the spatial proximity and overlapping nature of these classes in the image data. The 'Background clutter' class, by its very definition as a residual category, inevitably exhibits a high error rate due to its heterogeneous composition. A closer inspection of a specific image region (Figure~\ref{fig:example_inference_1}) offers further evidence of the model's limitations. In this particular instance, the challenge of distinguishing between 'Moving cars' and 'Static cars' is exacerbated by the static nature of the vehicles, which reduces the discriminative cues available to the model. The accurate segmentation of the 'Human' class is further hindered by the dense and overlapping nature of human figures within the scene. This visual comparison underscores the complexities inherent in accurately segmenting objects within crowded urban environments, particularly when dealing with overlapping instances, occlusions, and subtle visual distinctions between similar object categories.

Table~\ref{tab:results1_uavid} illustrates the effectiveness of the selected models in terms of the number of parameters, latency, and frames per second (FPS) for an image size of 1024x1024 pixels. SegFormer-B0 demonstrates satisfactory performance, achieving a $mIoU$ of 66.19\% with a latency of 7.67 milliseconds while utilizing only 3.7 million parameters. SegFormer-B0 model is well-suited for applications demanding real-time semantic segmentation of UAV images \cite{Safavi_2023}. For example, the model can continuously analyze frames from a camera mounted on a UAV, alerting for any potential hazards or dangerous situations it detects.

\begin{table}\centering
\caption{Models summary for a batch size of 1 and profiling on a NVIDIA A100-PCIe GPUs with 40GB memory and CUDA 11.5.}\label{tab:results1_uavid}
\scriptsize \begin{adjustbox}{width=0.7\linewidth}
\begin{tabular}{l|rrrr}\toprule
Model & Parameters & Image size & Latency (ms) & FPS  \\\midrule
SegFormer-B0 & 3.7M  & 1024x1024  &  7.67 & 132.16\\
SegFormer-B3   & 47.2M & 1024x1024  &  21.24 & 47.07 \\
SegFormer-B5    & 84.6M & 1024x1024 &  40.34 & 24.79 \\
\bottomrule
\end{tabular} \end{adjustbox}
\end{table}

\section{Conclusion}
In this study, we investigated the efficacy of SegFormer models in semantic segmentation tasks utilizing UAV images. Leveraging the SegFormer framework, tailored to accommodate various encoder sizes from B0 to B5, we conducted experiments on the UAVid dataset, specializing in urban scene semantic segmentation. Our investigation aimed to evaluate both the effectiveness and efficiency of SegFormer variants across different performance metrics. The results obtained shed light on the performance of SegFormer models across various encoder sizes. Notably, SegFormer models with larger encoders, such as MiX-B5, exhibited slightly superior performance compared to their counterparts with smaller encoders. However, it's important to highlight that the incorporation of test-time augmentation notably enhanced the overall performance of the models.

Furthermore, Ensemble methods emerged as a crucial strategy to further enhance the predictions of base SegFormer models. Combining predictions from multiple models, both with and without test-time augmentation, resulted in improved $mIoU$ values compared to the base models. This highlights the effectiveness of ensemble techniques in semantic segmentation tasks. While the SegFormer models demonstrated promising results across various labels, certain challenges were evident. The model tended to misclassify certain labels, such as "Moving car" and "Static car", which could be attributed to semantic similarities between these labels. Additionally, fine-grained segmentation for labels like "Human" proved challenging due to overlapping objects and dense segmentation in ground truth masks.

Despite these challenges, SegFormer-B0, the smallest model tailored for real-time applications, showcased satisfactory performance. With a mean $IoU$ of 66.187\% and a low latency of 7.67 milliseconds, while utilizing only 3.7 million parameters, SegFormer-B0 proves to be well-suited for real-time semantic segmentation of UAV images, offering promising prospects for practical applications in various domains, including environmental monitoring, disaster management, and aerial surveying. As further work, we aim to deploy this model on an edge device mounted on a UAV. This deployment would leverage the model's real-time capabilities, enhancing the UAV's ability to analyze and respond to its environment autonomously. This step is crucial for applications requiring immediate data processing and decision-making, ensuring timely alerts and responses to potential hazards or dangerous situations detected during UAV missions. In conclusion, this study underscores the effectiveness of SegFormer models in semantic segmentation tasks involving UAV images, providing valuable insights into their performance and potential applications in real-world scenarios.

\section{Acknowledgement}
The authors gratefully acknowledge the financial support provided by the Faculty of Computer Science and Engineering at the Ss. Cyril and Methodius University in Skopje through the SatTime project focused on satellite image time-series analysis.

\bibliographystyle{unsrt}  
\bibliography{references}  %%% Remove comment to use the external .bib file (using bibtex).
%%% and comment out the ``thebibliography'' section.

\end{document}